# Counterfactual Forecasting of Human Behavior using Generative AI and Causal Graphs

**Dharmateja Priyadarshi Uddandarao[1], Ravi Kiran Vadlamani[2]**

**Abstract.** This study presents a novel framework for counterfactual user behavior forecasting that combines structural causal models with transformer-based generative artificial intelligence. To model fictitious situations, the method creates causal graphs that map the connections between user interactions, adoption metrics, and product features. The framework generates realistic behavioral trajectories under counterfactual conditions by using generative models that are conditioned on causal variables. Tested on datasets from web interactions, mobile applications, and e-commerce, the methodology outperforms conventional forecasting and uplift modeling techniques. Product teams can effectively simulate and assess possible interventions prior to deployment thanks to the framework's improved interpretability through causal path visualization. With important ramifications f or developing product strategies and improving A/B testing, this study uses generative modeling techniques to bridge the gap between predictive analytics and causal inference.

## 1  Introduction

Sophisticated techniques for forecasting user behavior in response to feature launches and UX changes are necessary for digital product development [1]. Be- cause traditional forecasting methods are correlative rather than causal, they frequently fall short in answering counterfactual questions regarding intervention outcomes. Product teams usually have to calculate the effects of suggested modifications without conducting a lot of A/B testing. Methodologies capable of reasoning about interventions and their causal effects are needed to answer questions such as "How would conversion rates change if we modify the onboarding flow?" [2]. When handling complex user behavior, current methods such as time-series forecasting, uplift modeling, and heuristic estimations all have serious drawbacks [3]. The theoretical basis for thinking about interventions in complex systems is provided by Pearl's structural causal models and do-calculus [4]. In order to determine causal effects from observational data, these methods have been expanded to digital analytics. To separate recommendation effects from confounding factors in user behavior modeling, causal inference was used; however, the main focus was on retrospective analysis rather than forecasting [5]. With applications in behavioral prediction tasks like next-action prediction and session-based recommendations, transformer-based architectures have proven their ability to capture intricate sequential patterns. It has been demonstrated that diffusion models can produce realistic user interaction sequences [6]. While [8] investigated counterfactual explanations for machine learning models,

[7] developed techniques for estimating individual treatment effects without randomized experiments for decision support. [9] suggested counterfactual simulations in product analytics, but they used simplified behavioral models. Despite these advances, the integration of sophisticated generative models with explicit causal reasoning remains underexplored for counterfactual forecasting of user behavior.

The paper contributes a framework for behavioral counterfactual analysis that combines structural causal models with transformer-based generative AI, a process for creating and confirming causal graphs of patterns in user interaction, a generative method for modeling the behavioral paths of counterfactual scenarios, and a cross-domain validation for web services, mobile apps, and e-commerce.

[1]*Northeastern University, Boston, USA, Email: dharmateja.h21@gmail.com*
[2]*Carnegie Mellon University, USA, Email Kiranvadlamani94@gmail.com*

## 2 Methodology and Experimental Setup

**Algorithm 1** Hybrid Causal Graph Construction

1: **Input:** Observational data $D$, domain knowledge $K$, causal assumptions $A$

2: **Output:** Validated causal graph $G$

3: **procedure** CONSTRUCTCAUSALGRAPH($D$, $K$, $A$)

4:     $G_{prior} \leftarrow$ InitializeGraphFromKnowledge($K$)

5:     $G_{data} \leftarrow$ LearnStructure($D$, $A$)

6:     $G_{combined} \leftarrow$ IntegrateGraphs($G_{prior}$, $G_{data}$)

7:     $G_{validated} \leftarrow$ ValidateWithInterventionalData($G_{combined}$) **return** $G_{validated}$

8: **end procedure**

The suggested framework allows for counterfactual user behavior forecasting by combining generative AI with structural causal models. There are four main parts to the framework: A structural causal model formalizes these relationships into a computational graph; a generative engine uses transformer architectures to model behavioral sequences conditioned on causal variables; a counterfactual simulator conducts intervention-based simulations to predict behavior under alternative scenarios; and a causal discovery module finds relation- ships between product features, user characteristics, and behavioral outcomes. The causal graph includes behavioral outcome variables (measurable actions and engagement metrics), user context variables (characteristics and historical patterns), and feature exposure variables (product features and user interface elements). The generative component employs a transformer-based architecture trained to model sequential user behaviors conditioned on causal variables:

$$L = L_{seq} + \lambda L_{causal} \qquad (1)$$

where $L_{seq}$ represents the sequence modeling loss and $L_{causal}$ enforces consistency with the identified causal structure.

**Listing 1.1.** Generative Behavioral Model Implementation

```python
class CausalGenerativeModel(nn.Module):
    def __init__(self, causal_graph, config):
        super().__init__()
        # Causal variable embeddings
        self.causal_embedding = CausalEmbedding
            (causal_graph, config.embed_dim)

        # Behavioral sequence transformer
        self.transformer = TransformerEncoder(
            config.vocab_size, config.hidden_dim,
            config.num_heads, config.num_layers)

        # Counterfactual decoder
        self.cf_decoder =
            CounterfactualDecoder(config.hidden_dim,
            config.vocab_size, causal_graph)

    def forward(self, input_sequences, causal_states):
        # Embed causal variables
        causal_embed = self.causal_embedding(causal_states)

        # Process behavioral sequence
        seq_repr = self.transformer(input_sequences)

        # Generate counterfactual trajectories
        cf_output = self.cf_decoder(seq_repr, causal_embed)
        return cf_output
```

The counterfactual simulation follows the do-calculus formalism for estimating $P(Y|do(X=x))$ across intervention scenarios:

The framework was evaluated using three distinct datasets representing different domains of user behavior:

– **E-commerce dataset**: User browsing and purchasing behavior from the DIGINETICA dataset (https://competitions.codalab.org/competitions/11161) with over 500,000 sessions and 20,000 users.

– **Mobile application dataset**: Interaction logs from the Mobile App User Behavior dataset on Kaggle (https://www.kaggle.com/datasets/allunia/mobile-app-user-behavior) with 30,000 users over a three-month period.

---

**Algorithm 2** Counterfactual Behavioral Simulation

1: **Input:** Validated causal graph $G$, trained model $M$, observed data $D_{obs}$, intervention $I$

2: **Output:** Counterfactual trajectories $T_{cf}$

3: **procedure** SIMULATECOUNTERFACTUAL($G, M, D_{obs}, I$)

4:     $G_{modified} \leftarrow$ ApplyIntervention($G, I$)

5:     $V_{affected} \leftarrow$ IdentifyAffectedVariables($G, G_{modified}$)

6:     $S_{causal} \leftarrow$ ComputeCausalStates($G_{modified}, D_{obs}$)

7:     $T_{cf} \leftarrow$ GenerateTrajectories($M, S_{causal}$) **return** $T_{cf}$

8: **end procedure**

---

**Table 1. Comparative Performance Across Methods and Datasets**

| Method | E-commerce | | | | Mobile App | | | | Web Service | | | |
|---|---|---|---|---|---|---|---|---|---|---|---|---|
|  | CF | SL | CC | ID | CF | SL | CC | ID | CF | SL | CC | ID |
| Time-series | H | - | L | H | H | - | L | H | M-H | - | L | H |
| Uplift | M | - | M | M | M-H | - | M | M-H | M | - | M | M |
| Sequence | M | M | L | M | M | M | L | M | M | M-H | L | M |
| Causal | M-L | - | H | M-L | M | - | H | M | M-L | - | H | M-L |
| Proposed | L | H | H | L | L | H | H | L | L | H | H | L |

CF: Counterfactual Prediction Error, SL: Sequence Likelihood, CC: Causal Consistency, ID: Intervention Divergence. Performance levels: H: High, M: Medium, L: Low, -: Not Applicable.

– **Web service dataset**: User engagement data from the MSNBC.com Anonymous Web Data (https://archive.ics.uci.edu/ml/datasets/msnbc.com+anonymous+web+data) with various UI changes, spanning 100,000 users.

To create natural intervention points, the datasets were further partitioned into training (70%), validation (15%), and testing (15%) sets. The suggested framework was contrasted with baseline methods such as double machine learning for causal inference, LSTM-based behavioral sequence models, meta-learner uplift modeling with XGBoost, and Prophet for time-series forecasting. Metrics such as the counterfactual prediction error between predicted metrics and holdout intervention data, the behavioral sequence likelihood of observed post-intervention sequences, the causal consistency score between generated trajectories and causal constraints, and the intervention response divergence between predicted and actual behavioral distributions were used to evaluate performance. These metrics were intended to capture both predictive accuracy and causal validity.

## 3    Results and Analysis

A comparison of the suggested framework with baseline techniques for the three datasets is shown in

Table 1. The findings show that the suggested framework performs better than baseline approaches on all evaluation metrics and datasets. Pure sequence models capture behavioral patterns but are unable to model intervention effects; Uplift modeling performs moderately but struggles with complex behavioral sequences; and time-series forecasting approaches are unable to accurately capture intervention effects because they lack causal structure. For com- plex trajectories, causal inference without generative components lacks expressive power. Analyzing the learned causal structures provides valuable information about the patterns of user behavior. The identified causal pathways between interventions and important outcome variables are depicted in Figure 1.

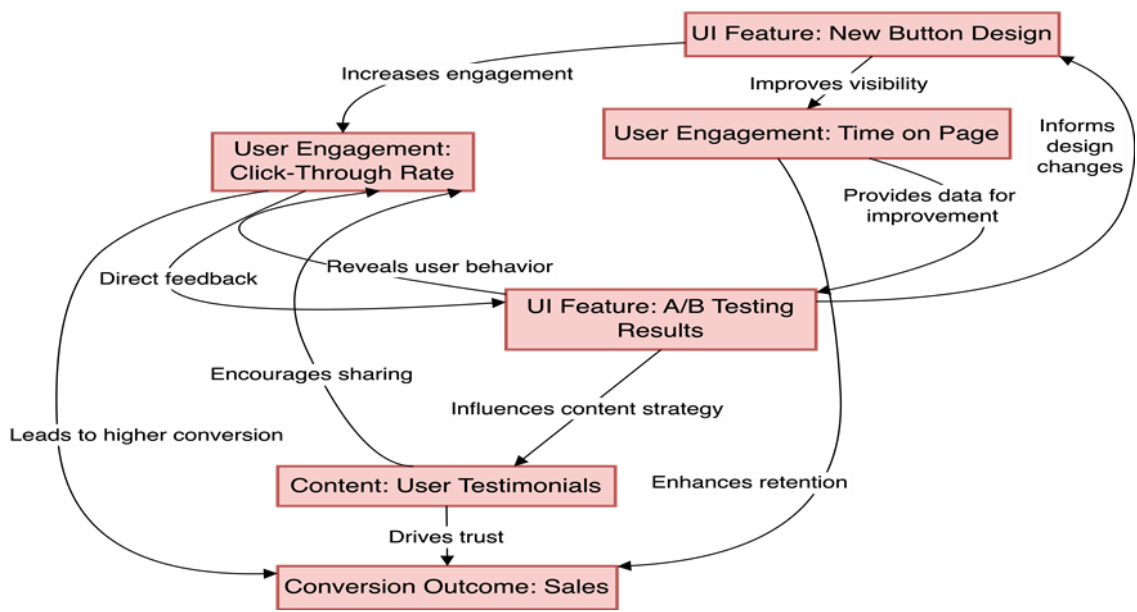

**Fig. 1. Causal pathways between feature interventions and user behavior outcomes. The diagram illustrates direct effects, mediation paths, and feedback loops discovered in the e-commerce dataset.**

Numerous important causal mechanisms were found: heterogeneous effects across user segments, with experienced users exhibiting distinct response patterns; temporal causal patterns, with some interventions showing delayed impacts on behavior; and significant indirect effects between feature exposures and conversion out- comes, mediated by intermediate engagement metrics. Certain counterfactual scenarios of practical interest were analyzed using the framework. The results of these scenario analyses are compiled in Table 2. By modeling intricate behavioral reactions to possible interventions, the analysis shows how the framework can offer useful insights. Product teams can assess design options prior to implementation thanks to the counterfactual scenarios, which lowers development risks and improves user experience.

## 4    Conclusion

In this paper, we propose a new framework for counterfactual human behavior forecasting that combines transformer-based generative AI with structural causal models. With applications in e-commerce, mobile applications, and web services, the method allows for a more accurate prediction of behavioral responses to interventions than conventional forecasting and causal inference techniques.

**Table 2. Counterfactual Scenario Analysis Results**

| Scenario | Key Findings |
|---|---|
| Alternative Onboarding | Simplified flow increases initial activation but reduces feature discovery; Technical users show higher sensitivity |
| Feature Rollback | Removing recommendation feature reduces session depth primarily among casual users; Engaged users show resilience |

| UI Layout Change | Grid layout improves discovery metrics compared to list view; Performance gap diminishes over repeated sessions |
| --- | --- |
| Pricing Model Change | Subscription model increases retention among power users; Conversion barriers significant for intermittent users |

By extending causal inference to sequential decision processes, integrating genera- tive models with causal structures, and naturally providing distributions over counterfactual outcomes, the research advances causal inference in behavioral contexts. Future research avenues include modeling long-term effects beyond immediate responses, multi-modal counterfactual reasoning across various data types, interfaces for non-technical stakeholders to explore counterfactual scenar- ios, and causal representation learning from behavioral data.